# A Classification–Based Study of Covariate Shift in GAN Distributions

Shibani Santurkar [1]  Ludwig Schmidt [1]  Aleksander Madry [1]

## Abstract

A basic, and still largely unanswered, question in the context of Generative Adversarial Networks (GANs) is whether they are truly able to capture all the fundamental characteristics of the distributions they are trained on. In particular, evaluating the diversity of GAN distributions is challenging and existing methods provide only a partial understanding of this issue. In this paper, we develop quantitative and scalable tools for assessing the diversity of GAN distributions. Specifically, we take a classification-based perspective and view loss of diversity as a form of covariate shift introduced by GANs. We examine two specific forms of such shift: mode collapse and boundary distortion. In contrast to prior work, our methods need only minimal human supervision and can be readily applied to state-of-the-art GANs on large, canonical datasets. Examining popular GANs using our tools indicates that these GANs have significant problems in reproducing the more distributional properties of their training dataset.

## 1. Introduction

Generative Adversarial Networks (Goodfellow et al., 2014) have achieved impressive results in producing realistic samples of natural images. They thus have become a promising approach to learning generative models. How well can GANs learn the truly underlying data distribution though? Answering this question would be key to properly understanding the power and limitations of the GAN framework, and the effectiveness of the adversarial setup.

A natural first step to evaluate the distribution-learning performance of a GAN is to examine whether the generated samples lie in the support of the true distribution, i.e., the distribution the GAN was trained on. In the case of images - arguably the most common application domain for GANs - this corresponds to checking if the GAN samples look realistic and are of good quality. Visual inspection of generated images by a human is currently the most widespread way of performing such checks, and from this perspective they indeed achieve impressive results (Denton et al., 2015; Karras et al., 2017). These successes were further corroborated by metrics such as the Inception Score (Salimans et al., 2016) and other divergence measures (Im et al., 2018).

[1]Massachusetts Institute of Technology. Correspondence to: Shibani Santurkar <shibani@mit.edu>, Ludwig Schmidt <ludwigs@mit.edu>, Aleksander Madry <madry@mit.edu>.

Once we have established that GANs produce images that look realistic, the next concern might be that this is due to them simply memorizing the training data. While this hypothesis cannot be ruled out entirely, there are reasons to believe that GANs indeed tend to avoid this deficiency. In particular, previous studies show that interpolations in the latent space of the generator produce novel image variations (Radford et al., 2015), and that there is a clear distinction between GAN samples and their nearest neighbors in the true dataset (Arora & Zhang, 2017). Taken together, these results provide evidence that GANs, at the very least, perform some non-trivial version of learning the true distribution. The question that remains though is: To what extent do GANs capture the *full* diversity of the underlying true distribution?

Prior work (Arora & Zhang, 2017; Goodfellow, 2016) has provided a methodology to study the diversity of the distributions that GANs learn. However these approaches fall short of examining the full extent of the potential problems. This is so as they tend to consider only simplified models (Goodfellow, 2016; Metz et al., 2016), rely heavily on manual annotation (Arora & Zhang, 2017) or fail to detect certain basic forms of diversity loss (Salimans et al., 2016).

The focus of this work is thus to address these shortcomings and pursue the following question:

*Can we develop a quantitative and universal methodology for studying diversity in GAN distributions?*

The approach we take uses classification as a lens to examine the diversity of GAN distributions. More precisely, we aim to measure the covariate shift (Sugiyama et al., 2006) with respect to the true distribution that GANs introduce. The key motivation here is that were GANs able to fully learn the true distribution, they would exhibit no such shift.

Specifically:

- We propose a framework to measure covariate shift introduced by GANs from a classification perspective.

- We demonstrate two specific forms of covariate shift caused by GANs: 1) Mode collapse, which has been observed in prior work (Goodfellow, 2016; Metz et al., 2016); 2) Boundary distortion, a phenomenon identified in this work and corresponding to a drop in diversity of the periphery of the learned distribution.

- Our methods need minimal human supervision and can easily be scaled to evaluating state-of-the-art GANs on the same datasets they are typically trained on.

To demonstrate the effectiveness of our approach, we chose five popular GANs and studied them on the CelebA and LSUN datasets – arguably the two most well known datasets in the context of GANs. Interestingly, we find that all the studied adversarial setups suffer significantly from the types of diversity loss we consider.

### 1.1. Related Work

The most direct approach to evaluate the performance of a GAN is to measure the difference between the probability density of the generator and the true distribution. One technique to do this is to quantify the log-likelihood of true test data under a given generative model. Various log-likelihood estimates have been developed using tools such as Parzen windows (Theis et al., 2016) and annealed importance sampling (Wu et al., 2016). However, in the context of GANs these approaches face two difficulties. First, GANs do not provide probability estimates for their samples. Second, GANs are not equipped with a decoder functionality, which makes log-likelihood estimation intractable. Hence, prior work was able to estimate the learned density only on simple datasets such as MNIST (Wu et al., 2016). Alternatively, researchers have studied simplified models such as GANs for two-dimensional Gaussian mixtures. These studies were insightful. For instance, they showed that one of the most common failures of GAN distributions is loss of diversity via *mode collapse* (Goodfellow, 2016; Metz et al., 2016). However, due to the simplicity of these models, it is unclear how exactly these observations translate to state-of-the-art GAN setting.

In another line of work, researches have developed methods to score overall GAN performance (Salimans et al., 2016; Che et al., 2016; Heusel et al., 2017; Im et al., 2018). Salimans et al. (2016) propose the Inception Score, which uses entropy in labels output by Inception network to assess similarity between true and GAN samples. This metric however has known shortcomings – for instance, a model can get a high Inception score even if it collapses to a single image (Che et al., 2016; Heusel et al., 2017). Che et al. (2016) develop a combined metric of visual quality and variety known as the MODE score. The authors however explicitly avoid using it on the CelebA and LSUN datasets (typical GAN datasets, used in our evaluations). Arora & Zhang (2017) propose an approach to directly get a handle on the diversity in GANs by leveraging the birthday paradox. They conclude that the learned distributions in the studied GANs indeed have (moderately) low support size. This approach however heavily relies on human annotation. Hence it does not scale easily or enable asking more fine-grained questions than basic collision statistics.

## 2. Understanding GANs through the Lens of Classification – Basic Illustration

Our goal is to understand the diversity of the distributions learned by GANs from the perspective of classification. In particular, we want to view the loss of diversity in GAN distributions as a form of covariate shift with respect to the true distribution. We focus on two forms of covariate shift that GANs could introduce: mode collapse and boundary distortion. We first develop intuition for these phenomena by discussing them in the context of simple Gaussian distributions. We then describe, in Section 3, a setup to precisely measure them in general GANs.

### 2.1. Gaussian Setting

Prior studies on learning two-dimensional multi-modal Gaussians with GANs helped to identify the phenomenon of mode collapse (Goodfellow, 2016; Metz et al., 2016). This phenomenon can be viewed as a form of covariate shift in GANs wherein the generator concentrates a large probability mass on a few modes of the true distribution. In Section 3.1, we augment this understanding by describing how this effect can be measured in general GANs using classification-based methods. In particular, we provide a way to obtain a more fine-grained understanding of mode collapse that goes beyond the simple Gaussian setting.

Now, to illustrate boundary distortion, we consider the simplest possible setup: learning a unimodal spherical Gaussian distribution. The key here is to understand how well can a GAN capture this data distribution. In Figure 1(a), we show an example of learning such a distribution using a vanilla GAN (Goodfellow et al., 2014). Shown in the figure is the spectrum (eigenvalues of the covariance matrix) for true data and GAN data. We observe that the spectrum of GAN data has a decaying behavior. (This observation was consistent across multiple similar experiments.) It is also clear that a distribution with such a skewed spectrum would indeed not properly capture diversity in the boundary regions of the support. In Figure 1(b), we show how such boundary distortion could cause errors in classification. In this illustration, we consider binary classification using lo-

# A Classification–Based Study of Covariate Shift in GAN Distributions

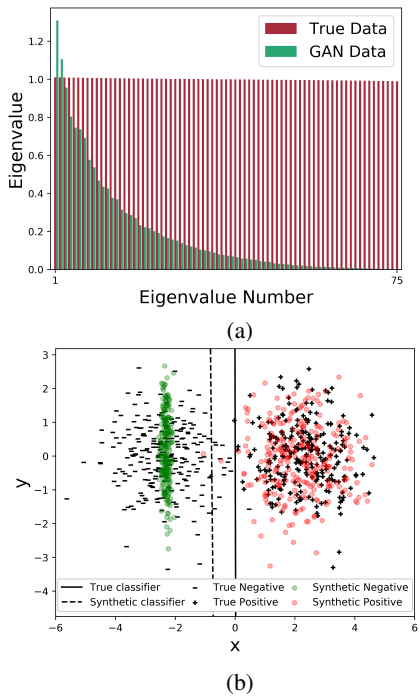

*Figure 1.* (a) Spectrum of the learned distribution of a vanilla GAN, compared to that of the true distribution. The true distribution is a 75-dimensional spherical unimodal Gaussian, while the GAN latent space is 200-dimensional (setup and details are in the Appendix). (b) An example of error in the learned decision boundary of a (linear) logistic regression classifier under covariate shift between synthetic and true distributions. Here the synthetic distribution for one class suffers from boundary distortion.

gistic regression where the true distribution for each class is a unimodal spherical Gaussian and the synthetic distribution for one of the classes undergoes boundary distortion. It is evident that this phenomenon causes a skew between the classifiers trained on true and synthetic data. Naturally, such errors would lead to poor generalization performance on true data as well. Taken together, these visualizations show that: boundary distortion is a form of covariate shift that GANs can realistically introduce; and this form of diversity loss can be detected and quantified even using classification.

### 2.2. Defining Covariate Shift in GANs

We will be interested in studying covariate shift between the true distribution, $P_T(\mathbf{X})$, and the GAN distribution, $P_G(\mathbf{X})$. This occurs when $P_T(\mathbf{Y}|\mathbf{X}) = P_G(\mathbf{Y}|\mathbf{X})$, but $P_T(\mathbf{X}) \neq P_G(\mathbf{X})$, for a source domain $\mathbf{X}$ (images) and target domain $\mathbf{Y}$ (classes) (Quionero-Candela et al., 2009). This can happen even when the GAN produces good quality images with valid class labels (matching the true distribution), as long as it, for instance, distorts the marginal distribution over the image space ($P_G(\mathbf{X})$). If a GAN successfully models the true data distribution, then naturally $P_T(\mathbf{X}, \mathbf{Y}) = P_G(\mathbf{X}, \mathbf{Y})$. Hence a GAN that perfectly learns the true distribution would introduce no such covariate shift.

## 3. Understanding GANs through the Lens of Classification – A General Framework

In Section 2.1, we used a simple Gaussian setting to illustrate two forms of covariate shift that GANs could introduce. While such studies are helpful, they are not sufficient to understand the behavior of GANs on more complex datasets. Therefore, in the following sections, we describe a quantitative and scalable methodology to capture and measure these forms of covariate shift in general GANs.

### 3.1. Mode Collapse

Our aim is to develop a more fine-grained and general understanding of mode collapse. As discussed previously, studies of this form have been so far largely restricted to simple Gaussian models or heavily reliant on human annotation.

Our point of start is a very basic question: if a GAN is trained on a well-balanced dataset (i.e., a dataset that contains equal number of samples from each class), can it learn to reproduce this (simple) balanced structure?

To answer this question for a particular GAN, we propose the following approach:

1. Train the GAN unconditionally (without class labels) on the chosen balanced multi-class dataset $D$.

2. Train a multi-class classifier on the same dataset $D$.

3. Generate a synthetic dataset by sampling $N$ images from the GAN. Then use the classifier trained in Step 2 above to obtain labels for this synthetic dataset.

One can think of the classifier trained in Step 2 as a scalable 'annotator', thereby circumventing the laborious manual annotation needed in prior work. We can now use the labels produced by this annotator to obtain a quantitative assessment of the mode distribution learned by GANs.

A visualization of the mode distribution in five popular GANs is shown in Figure 2 (setup details can be found in Section 5.1). It is evident that all the studied GAN exhibit prominent mode collapse and they do not recover from this effect as training progresses. Note that our approach is not only more scalable as compared to prior work, but also allows for asking deeper and more fine-grained questions on the extent of mode collapse. To do this one just needs to control the difficulty of the classification tasks, i.e., choose suitably challenging classes or vary their number.

### 3.2. Boundary Distortion

In Section 2.1, we illustrated boundary distortion caused by GANs and how this effect can be detected by leveraging the classification perspective. To extend this analysis to more complex distributions, we recall a key observation: classifiers trained on a synthetic distribution with boundary distortion will likely generalize poorly to the true distribution (Section 2.1). This insight is grounded in prior studies of covariate shift (Sugiyama et al., 2006; Quionero-Candela et al., 2009). This line of research has demonstrated that in the presence of covariate shift between the training and test distributions, a classifier trained using empirical risk minimization will likely not be optimal (with respect to the test distribution) even asymptotically. In the context of GANs, this means that a classifier trained on $P_G(\mathbf{Y}|\mathbf{X})$ will not generalize to $P_T(\mathbf{Y}|\mathbf{X})$ in the presence of such shift (Section 2.2).

This motivates us to consider the question: how well can a GAN-generated dataset perform on a general binary classification task? Performance here refers to accuracy on a hold-out set of true data. As discussed above, this would measure the boundary distortion introduced by the GAN. With this in mind, we propose the following approach to measure such distortion caused by a given GAN:

1. Train two separate instances of the given unconditional GAN, one for each class in true dataset $D$.
2. Generate a balanced dataset by drawing $N/2$ from each of these GANs. Labels (which we refer to as the "default" labels) for this dataset are trivially determined from the class modeled by the GAN a given sample was drawn from.
3. Train a binary classifier based on the labeled GAN dataset obtained in Step 2 above.
4. Train an identical (in terms of architecture and hyperparameters) classifier on the true data $D$ for comparison.

Using this framework, we can investigate the boundary distortion caused by a given GAN by asking two questions:

(i) How easily can the classifier fit the training data? This reflects the complexity of the decision boundary in the training dataset.

(ii) How well does a classifier trained on synthetic data generalize to the true distribution? This acts as a proxy measure for diversity loss through covariate shift.

Note that the training of the GAN under study does not depend on the classification task that will be used for evaluation. Thus, obtaining good generalization performance indicates that the GAN likely reconstructs the *whole* boundary with high fidelity.

In our experiments based on this setup (Section 5.2), we observe that all the studied GANs suffer from significant boundary distortion. In particular, the accuracy achieved by a classifier trained on GAN data is comparable to the accuracy of a classifier trained on a $100\times$ (or more) subsampled version of the true dataset.

In this work, we perform this analysis based on binary classification primarily because it appears sufficient to illustrate issues with the learned diversity in GANs. This is by no means a restriction of our approach – it is straightforward to extend this setup to perform more complex multi-class tests on the GAN data, if needed. We would also like to point to a similar study conducted by Radford et al. (2015) for the *conditional* DCGAN on the MNIST dataset. They observe that the GAN data attains good generalization performance (we also could reproduce this result). However, we believe that this setting might be too simple to detect diversity issues-it is known that a classifier can get good test accuracy on MNIST even with a small training set (Rolnick et al., 2017). We thus focus on studying GANs in more complex settings-such as the CelebA and LSUN datasets (Section 5.2).

## 4. Experimental Setup

In the following sections we describe the setup and results for our classification-based GAN diversity studies. Additional details can be found in the Appendix.

### 4.1. Datasets

GANs have been applied to a broad spectrum of datasets. However, the CelebA (Liu et al., 2015) and LSUN (Yu et al., 2015) datasets remain the most popular and canonical ones in the context of their development and evaluation. Conveniently, these datasets also have rich annotations, making them particularly suited for our classification–based evaluations. We also explored the CIFAR-10 dataset (Krizhevsky & Hinton, 2009), however, the standard implementations of many of the studied GANs failed to produce meaningful images in this setting. Hence, our focus was on datasets that most GANs have been designed and optimized for, and that also have been the object of related work (Arora & Zhang, 2017). Image samples and details of the setup for our classification tasks are in the Appendix.

### 4.2. Models

Using our framework, we perform a comparative study of five popular variants of GANs: (1) DCGAN (Radford et al., 2015) (2) WGAN (Arjovsky et al., 2017) (3) ALI (Dumoulin et al., 2017) (4) BEGAN (Berthelot et al., 2017) and (5) Improved GAN or ImGAN (Salimans et al., 2016). This selection was motivated by: wanting to study learned diversity in a wide range of classic adversarial setups (different distance measures such as Jensen-Shanon Divergence and Wasserstein distance, and training with joint inference mechanism); and studying the performance of some newer variants such as BEGAN. We use standard implementations and prescribed hyper-parameter settings for each of these models, details of which are provided in the Appendix. BEGAN did not converge in our experiments on the LSUN dataset and hence is excluded from the corresponding analysis.

In our study, we use two types of classification models:
1. 32-Layer ResNet (He et al., 2016): We choose a ResNet as it is a standard classifier in vision and yields high accuracy on various datasets, making it a reliable baseline.
2. Linear Model: This is a network with one-fully connected layer between the input and output with a softmax non-linearity. Due to it's simplicity, this model will serve as a useful baseline in some of our experiments.

The same architecture and hyperparameter settings were used for all datasets (true and GAN derived) in any given comparison of classification performance.

## 5. Results of Covariate Shift Analysis

We now describe results from our experimental studies of covariate shift in GANs based on the procedures outlined in Section 3, using the setup from Section 4.

### 5.1. Results of Mode Collapse Analysis

Figure 2 presents class distribution in synthetic data, as determined by the annotator classifiers. The left panel compares the relative distribution of modes in true data (uniform) with that in various GAN-generated datasets. The right panel illustrates the evolution of class distributions in various GANs over the course of training[1]. We present details of the datasets used in this analysis - such as size, number of classes, and annotator accuracy - in Appendix Table 2.

**Results:** These visualization lead to the following findings:

- We observe that all GANs suffer from significant mode-collapse. This becomes more apparent when the annotator granularity is increased, by considering a larger set of classes. For instance, one should compare the relatively balanced class distributions in the 3-class LSUN task to the near-absence of some modes in the 5-class task.

- Mode collapse is prevalent in GANs throughout the training process, and does not seem to recede over time. Instead the dominant mode(s) often fluctuate wildly over the course of the training.

- For each task, often there is a common set of modes onto which the distint GANs exhibit collapse.

Thus, such analysis enables us to elevate the discussion of mode collapse in high-dimensional distributions to one which can be quantified, visualized and compared across various GANs.

From this perspective, we observe that the ALI setup consistently shows lowest mode collapse amongst all the studied GANs. The behavior of the other GANs appears to vary based on the dataset – on CelebA, DCGAN learns a somewhat balanced distributions, while WGAN, BEGAN and ImGAN show prominent mode collapse. This is in contrast to the results obtained LSUN, where, WGAN exhibit relatively small mode collapse, while DCGAN and ImGAN show significant mode collapse. This points to a general challenge in real world applications of GANs: they often perform well on the datasets they were designed for (e.g. WGAN on LSUN), but extension to new datasets is not straightforward. Temporal analysis of mode-collapse shows that there is wide variation in the dominant mode for WGAN and ImGAN, whereas for BEGAN, the same mode(s) dominates the entire training process.

### 5.2. Boundary Distortion Analysis

Selected results for classification-based evaluation of boundary distortion caused by GANs are shown in Table 1. Additional results are in Appendix Figures 3, 5 and 6.

As a preliminary check, we verify that the covariate shift assumption from Section 3, i.e. $P_T(\mathbf{Y}|\mathbf{X}) = P_G(\mathbf{Y}|\mathbf{X})$ holds. For this, we use high-accuracy *annotator* classifiers (Section 3.1) to predict labels for GAN generated data and measure consistency between the predicted and default labels (label correctness). We also inspect confidence scores of the annotator, which is defined as the softmax probabilities for predicted class. Another motivation for these measures is that if an annotator can correctly and with high-confidence predict labels for labeled GAN samples, then it is likely that they are convincing examples of that class, and hence of good "quality". This would suggest that any drop in classification performance from training on synthetic data was more likely due to sample "diversity" issues.

It is clear from Table 1, that the label correctness is high for GAN datasets and is on par with the scores for true test data. These checks affirm that the covariate shift assumption holds and that GAN images are of high-quality, as expected based on the visual inspection. These scores are slightly lower for LSUN, potentially due to lower quality of LSUN GAN images.

In Table 1, we also report a modified Inception Score (Salimans et al., 2016). The standard Inception Score uses labels from an Inception network trained on ImageNet. While these statistics might be meaningful for similar datasets studied by the authors such as CIFAR-10, it is unclear how informative this label distribution is for face or scene datasets, where all images belong to the same high-level category. We compute a modified Inception Score, based on the labels predicted by the annotator networks (trained on true data from the same distribution) instead of using a pre-trained Inception Network. Then the score is computed exactly like the Inception Score using $exp(\mathbb{E}_x[\mathbf{KL}(p(y|x))||p(y)])$, where $y$ is the label prediction.

---
[1]Temporal evaluation of ALI class distribution is absent in the analysis due to absence of periodic checkpointing in the code.

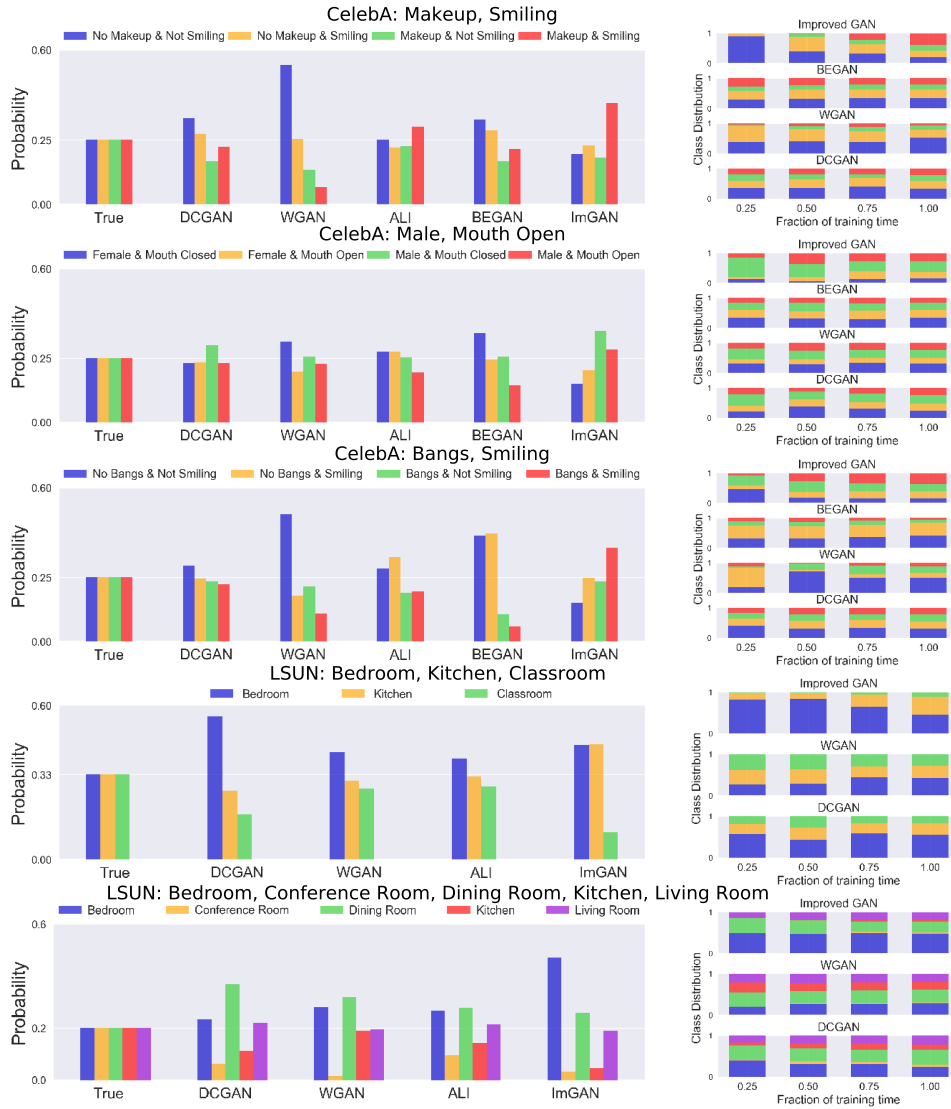

*Figure 2.* Illustration of mode collapse in GANs trained on select subsets of CelebA and LSUN datasets (Section 5.1). Left panel shows the relative distribution of modes in samples drawn from the GANs, and compares is to the true data distribution (leftmost plots). On the right, shown is the evolution of class distributions in different GANs over the course of training. BEGAN did not converge for LSUN and hence is excluded from the corresponding analysis. It is clear that all these GANs introduce covariate shift through mode collapse.

Next, to study boundary distortion in GANs, we measure the relative performance of a classifier trained on true and synthetic GAN datasets, in terms of test accuracy on true data. ResNets (and other deep variants) get good test accuracy when trained on true data, but suffer from severe overfitting on synthetic GAN data. This already indicates a possible problem with GANs and the diversity of the data they generate. But to highlight this problem better and avoid the issues that stem from overfitting, we also report performance for a classifier which does not always overfit on the synthetic data. We observed that even simple networks, such as a few fully connected layers, overfit to synthetic data. Hence, we selected a very basic *linear* model described in Section 4.2. Table 1 and Appendix Table 3 shows results from binary classification experiments using both deep ResNets and linear models for various datasets. Note that for a particular column, the same architecture and hyperparameters were used for classification using every true/synthetic dataset.

Finally, to get a better understanding of the extent of boundary distortion in GAN datasets, we train linear models using down-sampled versions of true data (no augmentation)

**A Classification–Based Study of Covariate Shift in GAN Distributions**

| Task | Classification Performance | | | | | | |
|---|---|---|---|---|---|---|---|
| | Data Source | Label Correctness (%) | Inception Score ($\mu \pm \sigma$) | Accuracy (%) | | | |
| | | | | Linear model | | | ResNet |
| | | | | $\uparrow_1$ | | $\uparrow_{10}$ | $\uparrow_1$ |
| | | | | Train | Test | Train | Test | Test |
| CelebA Smiling (Y/N) # Images: 156160 | True | | | 85.7 | 85.6 | | | 92.4 |
| | True $\downarrow_{64}$ | | | 87.6 | 85.0 | | | 87.8 |
| | True $\downarrow_{256}$ | 92.4 | $1.69 \pm 0.0074$ | 91.5 | 82.4 | | | 82.1 |
| | True $\downarrow_{512}$ | | | 93.7 | 80.0 | | | 77.8 |
| | True $\downarrow_{1024}$ | | | 95.0 | 76.2 | | | 71.2 |
| | DCGAN | 96.1 | $1.67 \pm 0.0028$ | 100.0 | 77.1 | 100.0 | 77.1 | 63.3 |
| | WGAN | 98.2 | $1.68 \pm 0.0031$ | 96.8 | 83.4 | 96.8 | 83.5 | 65.3 |
| | ALI | 93.3 | $1.71 \pm 0.0027$ | 94.5 | 80.1 | 95.0 | 82.4 | 55.8 |
| | BEGAN | 93.5 | $1.74 \pm 0.0028$ | 98.5 | 69.5 | 98.5 | 69.6 | 64.1 |
| | Improved GAN | 98.4 | $1.88 \pm 0.0021$ | 100.0 | 70.2 | 100.0 | 70.1 | 61.6 |
| LSUN Bedroom/Kitchen # Images: 200000 | True | | | 64.7 | 64.1 | | | 99.1 |
| | True$\downarrow_{512}$ | | | 64.7 | 64.0 | | | 76.4 |
| | True $\downarrow_{1024}$ | 98.2 | $1.94 \pm 0.0217$ | 65.2 | 64.0 | | | 66.9 |
| | True $\downarrow_{2048}$ | | | 98.7 | 56.2 | | | 56.5 |
| | True $\downarrow_{4096}$ | | | 100.0 | 55.1 | | | 55.1 |
| | DCGAN | 92.7 | $1.85 \pm 0.0036$ | 90.8 | 56.5 | 91.2 | 56.3 | 51.2 |
| | WGAN | 87.8 | $1.70 \pm 0.0023$ | 86.2 | 58.2 | 96.3 | 54.1 | 55.7 |
| | ALI | 94.1 | $1.86 \pm 0.0021$ | 82.5 | 60.0 | 82.0 | 59.7 | 56.2 |
| | Improved GAN | 84.2 | $1.68 \pm 0.0030$ | 91.6 | 55.9 | 90.8 | 56.5 | 51.2 |

*Table 1.* Select results measuring a specific form of covariate shift that GANs could introduce, namely boundary distortion, on the CelebA and LSUN datasets. Label correctness measures the agreement between default labels for the synthetic datasets, and those predicted by the annotator, a classifier trained on true data. Shown alongside are the modified inception scores computed using labels predicted by the annotator (rather than an Inception Network). Training and test accuracies for a linear model on the various true and synthetic datasets are reported. Also presented are the corresponding accuracies for this classifier trained on down-sampled true data ($\downarrow_M$) and oversampled synthetic data ($\uparrow_L$). Test accuracy for ResNets trained on these datasets is also shown (training accuracy was always 100%). These results show that all the studied GANs suffer significantly from diversity loss, especially near the periphery of the support.

for comparison (Table 1 and Appendix Table 3). Down-sampling the data by a factor of $M$, denoted as $\downarrow_M$ implies selecting a random $N/M$ subset of the data. Visualizations of how GAN classification performance compares with (down-sampled) true data are in Appendix Figures 6. An argument in the defense of GANs is that we can oversample them, i.e. generate datasets much larger than the size of training data. Results for linear models trained using a 10-fold oversampling of GANs (drawing $10N$ samples), denoted by $\uparrow_{10}$, are show in Tables 1 and Appendix Table 3.

**Results:** Based on experimental results, shown in Table 1, Appendix Figures 3 and 5, our major findings are:

- There is a large gap between generalization performance of classifiers trained on true and synthetic datasets. Given the high scores of synthetic data on the previous checks of dataset *quality*, it is likely that the poor classification performance is more indicative of lack of *diversity*.

- Inspection of the training accuracy shows that linear models are able to nearly fit the GAN datasets, but are grossly underfitting on true data. This would indicate that the decision boundary for GAN data is significantly less complex (probably due to an inability to full capture diversity near the boundary).

- Comparing GAN performance to that of down-sampled true data reveals that the learned distribution, which was trained on datasets with *around hundred thousand* data points exhibits performance that is on par with that of *only mere couple of hundreds* of true data samples! Further, oversampling GANs by 10-fold to produce larger datasets does not improve classification performance. The disparity between true and synthetic data remains nearly unchanged even after this significant oversampling, further highlighting issues with the learned diversity in GANs.

These results strongly indicate that, from the viewpoint of classification, there is a clear covariate shift between the true and GAN distributions through boundary distortion. Note that the modified Inception Score is not very informative and is similar for the true (test) and GAN datasets. This is not surprising as it has been suggested that a model could get a high Inception Score even if it collapses on a single image (Hendrycks & Basart, 2017; Che et al., 2016).

Amongst the studied GANs, WGAN and ALI seem to have highest diversity as per our measures. Surprisingly, while BEGAN samples have good perceptual quality, they consistently perform badly on our classification tasks. This is a strong indicator of the need to consider other metrics, such as the ones proposed in this paper, in addition to visual inspection to study GANs. For LSUN, the gap between true and synthetic data is larger, with the classifiers getting near random performance on many GAN datasets. Note that these classifiers get poor test accuracy on LSUN but are not overfitting on the training data. In this case, we speculate the lower performance could be due to both lower quality and diversity (boundary distortion) of LSUN samples.

Note that in this work, our aim was not to provide a relative ranking of state-of-the-art GANs, but instead to develop a framework that can allow for a fine-grained visualization of the learned diversity under various adversarial training setups. Thus the GANs studied in this paper are by no means exhaustive, and there have been several noteworthy developments such as IWGAN (Gulrajani et al., 2017), Progressive GAN (Karras et al., 2017) and MMD-GAN (Li et al., 2017) that we hope to analyze using our framework in the future. Our techniques could also, in principle, be applied to study other generative models such as Variational Auto-Encoders (VAEs) (Kingma & Welling, 2014). However, VAEs have known problems with generating realistic samples on the datasets used in our analysis (Arora & Zhang, 2017).

### 5.3. Discussion of GAN evaluation metrics

We believe that truly measuring the distribution learning capabilities of GANs, in particular the learned diversity, inherently requires using multiple metrics. Given that we aim to capture some very complex behaviour in terms of simple, concise metrics, each of these individually will inevitably conflate some aspects.

A GAN could, for instance, perform well on the proposed classification-based tests if it just memorized the training set. However, based on prior work, it is unlikely that state-of-the-art GANs have this deficiency as is discussed in Section 1.

Further, one might think that performance in the studies of boundary distortion caused by GANs (Section 3.2) would be largely based on diversity in the boundary region that is relevant for the specific classification task. While this may be true, recall that in our setup, the GAN is unaware of the classification task that will be used in the evaluation. Thus, obtaining good classification performance indicates that the GAN approximately reconstructs the whole boundary with high fidelity. Further, one could consider more complex multi-class tasks to develop a clearer understanding.

Our framework, in its current form, relies on the datasets having attribute labels. Still, we have a number of large-scale datasets with extensive annotation at our disposal (Deng et al., 2009; Yu et al., 2015; Liu et al., 2015) already. So, these datasets are able to provide a fairly comprehensive basis for evaluation of current generative models.

The proposed metrics are also somewhat dependent on the architecture of the classifier used. This is, however, true even for other GAN metrics: the Inception score (Salimans et al., 2016); and the discriminator based score (for CelebA and LSUN datasets) proposed in Che et al. (2016). Still, we believe that the relative performance of different GANs on these metrics, and how they compare to true data performance already provides meaningful information. In our experiments, we also studied other classifier networks and observed that the trends of the relative performance of the GANs was preserved irrespective of the choice of classifier.

Thus, we believe that simple metrics can capture only certain aspects of the quality or diversity in GAN distributions. Our metric should be used in conjunction with other evaluation approaches to develop a more wholesome understanding of the performance of GANs.

## 6. Conclusions

In this paper, we put forth techniques to examine the ability of GANs to capture key characteristics of the true distribution, through the lens of classification. In particular, we view the loss diversity in GAN distributions as a form of covariate shift the introduce. While our methodology is quite general, in this work, we use it to get a fine-grained understanding of two specific forms of such shift: mode collapse and boundary distortion. Our tools are scalable and quantitative, and thus be used to study and compare state-of-the-art GANs on large-scale image datasets.

We use our approach to perform empirical studies on popular GANs using the canonical CelebA and LSUN datasets. Our examination shows that mode collapse is indeed a prevalent issue for current GANs. Also, we observe that GAN-generated datasets have significantly reduced diversity in the periphery of the support. From a classification perspective, it appears that the diversity of GAN distributions is often comparable to true datasets that are a few orders of magnitude smaller. Furthermore, this gap in diversity does not seem to be bridged by simply producing much larger datasets by oversampling GANs. Finally, we also notice that good perceptual quality of samples does not necessarily correlate – and might sometime even anti-correlate – with distribution diversity. These findings suggest that we need to go beyond the visual inspection–based evaluations and look for different quantitative tools, such as the ones presented in this paper, to assess the quality of GANs.

## A. Experimental Setup

### A.1. Datasets for Classification Tasks

In the following subsections, we expand on details of the datasets using in our experimental analysis.

#### A.1.1. LOW-DIMENSIONAL GAUSSIAN DISTRIBUTION

The true distribution in 1(a), which shows boundary distortion in GANs, is a 75-dimensional, zero-mean unimodal spherical Gaussian distribution with an identity covariance matrix. In 1(b), which highlights the impact of covariate shift on the learnt decision boundary, the true distribution for each of the two classes consists of a two-dimensional unimodal spherical Gaussian. The synthetic distribution for the positive class is the same as the true distribution, while in the synthetic distribution for the negative class, the variance along the X-axis is reduced by two orders-of-magnitude.

#### A.1.2. IMAGE DATASETS

Our analysis is based on $64 \times 64$ samples of the CelebA and LSUN datasets, which is a size at which GAN generated samples tend to be of high quality. We also use visual inspection to ascertain that the perceptual quality of GAN samples in our experiments is comparable to those reported in previous studies as shown in Figures 3 and 4.

To assess GAN performance from the perspective of classification, we construct a set of classification tasks on the CelebA and LSUN datasets. In the case of the LSUN dataset, images are annotated with scene category labels, which makes it straightforward to use this data for binary and multi-class classification. On the other hand, each image in the CelebA dataset is labeled with 40 binary attributes. As a result, a single image has multiple associated attribute labels. Here, we construct classification tasks can by considering binary combinations of an attribute(s) (examples are shown in Figure 4).

Attributes used in our experiments were chosen such that the resulting dataset was large, and classifiers trained on true data got high-accuracy so as to be good *annotators* for the synthetic data. Details on datasets used in our classification tasks, such as training set size ($N$), number of classes ($C$), and accuracy of the annotator, i.e., a classifier pre-trained on true data which is used to label the synthetic GAN-generated data, are provided in Table 2.

### A.2. Models

Specifics of GAN and classifier models used in our experiments are described subsequently.

### A.3. Gaussian Classifier

For the Gaussian experiments described in 2.1, we use a simple vanilla GAN architecture [2]. We modify both the discriminator and the generator to have two-fully connected layers with ReLU non-linearities, followed by a final linear layer. The discriminator has a sigmoid activation layer after the final linear layer so as to scale outputs to be probabilities. The decision boundaries shown in 1(b) are based on a linear logistic regression classifier.

For the GAN distribution in 1(a), the error between the true sample mean, $\mu_T$, and the sample mean of the learned distribution in the GAN, $\mu_G$, is computed as $\delta_\mu = (\mu_T - \mu_G)^T \Sigma_T (\mu_T - \mu_G)$, where $\Sigma_T$ is the true sample covariance matrix. These estimates are obtained based on 2.5 million samples of each distribution.

### A.4. GANs

The high-dimensional benchmarks with common GANs were performed on standard implementations with recommended hyperparameter settings, including number of iterations we train them for -

1. Deep Convolutional GAN (DCGAN): Convolutional GAN trained using a Jensen–Shannon divergence–based objective (Radford et al., 2015); Code from https://github.com/carpedm20/DCGAN-tensorflow
2. Wasserstein GAN (WGAN): GAN that uses a Wasserstein distance–based objective (Arjovsky et al., 2017); Code from https://github.com/martinarjovsky/WassersteinGAN
3. Adversarially Learned Inference (ALI): GAN that uses joint adversarial training of generative and inference networks (Dumoulin et al., 2017); Code from https://github.com/IshmaelBelghazi/ALI
4. Boundary Equilibrium GAN (BEGAN): Auto-encoder style GAN trained using Wasserstein distance objective (Berthelot et al., 2017); Code from https://github.com/carpedm20/BEGAN-tensorflow
5. Improved GAN (ImGAN): GAN that uses semi-supervised learning (labels are part of GAN training), with various other architectural and procedural improvements (Salimans et al., 2016); Code from https://github.com/openai/improved-gan

All these GANs are unconditional, but ImGAN has access to class labels due to the semi-supervised training process.

---

[2] https://github.com/wiseodd/generative-models/blob/master/GAN/vanilla_gan/gan_tensorflow.py

| Dataset | $N$ | $C$ | Annotator's Accuracy (%) |
|---|---|---|---|
| CelebA: Makeup, Smiling | 102,436 | 4 | 90.9, 92.4 |
| CelebA: Male, Mouth Open | 115,660 | 4 | 97.9, 93.5 |
| CelebA: Bangs, Smiling | 45,196 | 4 | 93.9, 92.4 |
| LSUN: Bedroom, Kitchen, Classroom | 150,000 | 3 | 98.7 |
| LSUN: Bedroom, Conference Room, Dining Room, Kitchen, Living Room | 250,000 | 5 | 93.7 |

*Table 2.* Details of CelebA and LSUN subsets used for the studies in 5.1. Here, we use a classifier trained on true data as an *annotator* that let's us infer label distribution for the synthetic, GAN-generated data. $N$ is the size of the training set and $C$ is the number of classes in the true and synthetic datasets. Annotator's accuracy refers to the accuracy of the classifier on a test set of true data. For CelebA, we use a combination of attribute-wise binary classifiers as annotators due their higher accuracy compared to a single classifier trained jointly on all the four classes.

### A.5. Image Classifiers

The ResNet Classifier used was a variation of the standard TensorFlow ResNet. Code was obtained from https://github.com/tensorflow/models/blob/master/research/resnet/resnet_model.py.

The Linear Model used was a network with one-fully connected layer between the input and output with a softmax non-linearity. If the dimensions of input $x$ and output $\hat{y}$, are $D$ and $C$ respectively, then this linear model implements the function $\hat{y} = \sigma(W^T x + b)$, where $W$ is a $D \times C$ matrix, $b$ is a $C \times 1$ vector and $\sigma(\cdot)$ is the softmax function.

We train classifiers to convergence, with learning rate decay and no data augmentation.

The code used to compute Inception Score was based on a the original implementation provided by the authors from https://github.com/openai/improved-gan.

## B. Addition Experimental Results

### B.1. Gaussian Setting

In the experiment presented in Figure 1(a), the normalized error in means between true and GAN distributions is 0.35 and the KL Divergence is 2061.36. The generator and discriminator losses are 1.13 and 1.21 respectively.

### B.2. Sample Quality

For each of our benchmark experiments, we ascertain that the visual quality of samples produced by the GANs is comparable to that reported in prior work. Examples of random samples drawn for multi-class datasets from both true and synthetic data are shown in Figure 3 for the LSUN dataset and Figure 4 for the CelebA dataset.

### B.3. Boundary Distortion Experiments

3.2 and 5.2 describe techniques to study covariate shift due to boundary distortion introduced by GANs from a classification-based perspective. Table 3 presents an extension of the comparative study of classification performance of true and GAN generated data provided in 1. In order to assess the quality of the GAN-generated datasets, we evaluate label agreement between the default labels of the synthetic datasets, and that obtained using the *annotators* as shown in these tables. In Figure 5, we inspect the confidence scores, defined as the softmax probabilities for predicted class, of the annotator while making these predictions. Thus, it seems likely that the GAN generated samples are of good quality and are truly representative examples of their respective classes, as expected based on visual inspection.

Visualizations of how test accuracies of a linear model classifier trained on GAN data compares with one trained on true data is shown in Figure 6. For each task, the bold curve shows test accuracy of a classifier trained on true data as a function of true dataset size. A down-sampling factor of $M$ corresponds to training the classifier on a random $N/M$ subset of true data. The dashed curves show test accuracy of classifiers trained on GAN datasets, obtained by drawing $N$ samples from GANs at the culmination of the training process. Based on these visualizations, it is apparent that GANs have comparable classification performance to a subset of training data that is more than a 100x smaller.

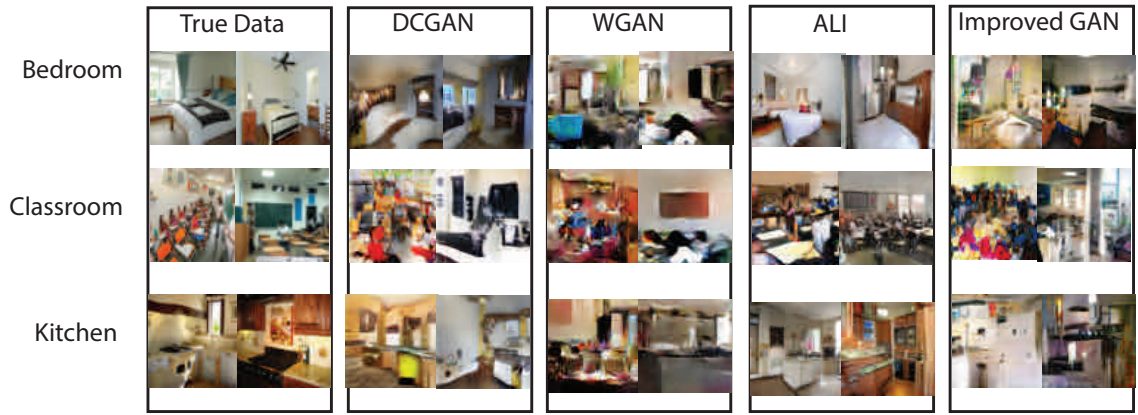

(a) 3-class dataset from LSUN for *Bedroom, Classroom, Kitchen*.

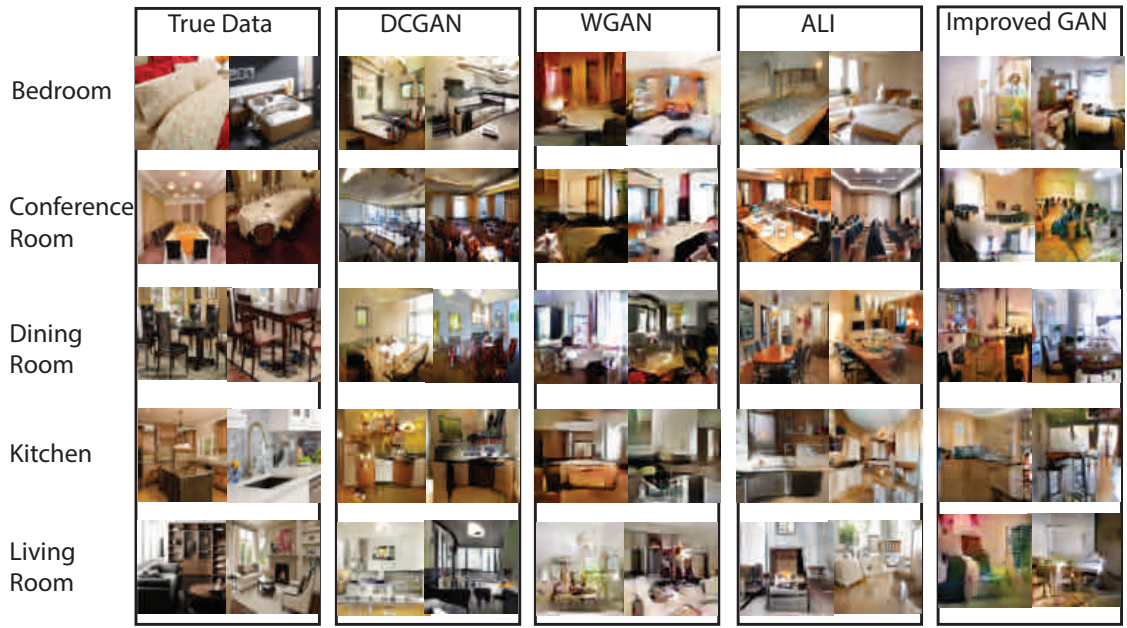

(b) 5-class dataset from LSUN for *Bedroom, Conference Room, Dining Room, Kitchen, Living Room*.

*Figure 3.* Illustration of subsets of the LSUN dataset used in our classification-based GAN studies. Shown alongside are samples from various GANs trained on this dataset. Labels for the GAN samples are obtained using a pre-trained classifier as an annotator.

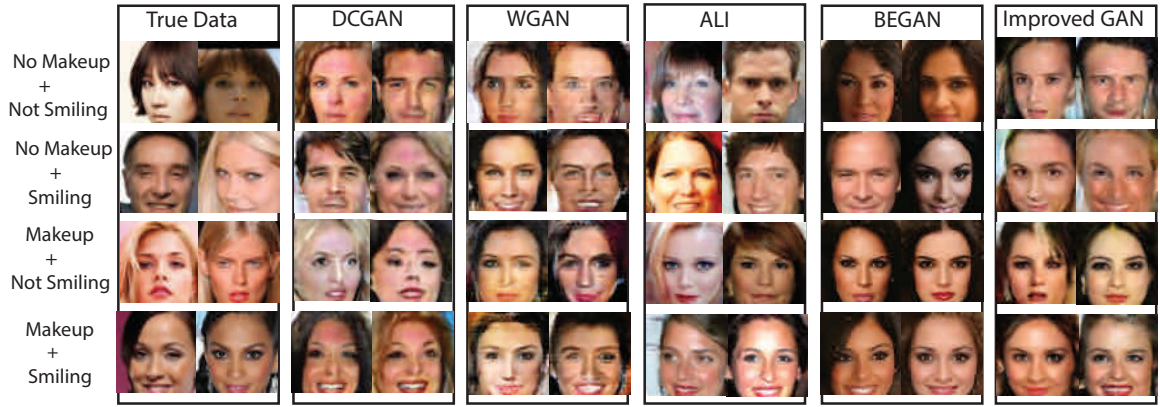

(a) 4-class dataset from CelebA for attributes *Makeup, Smiling*.

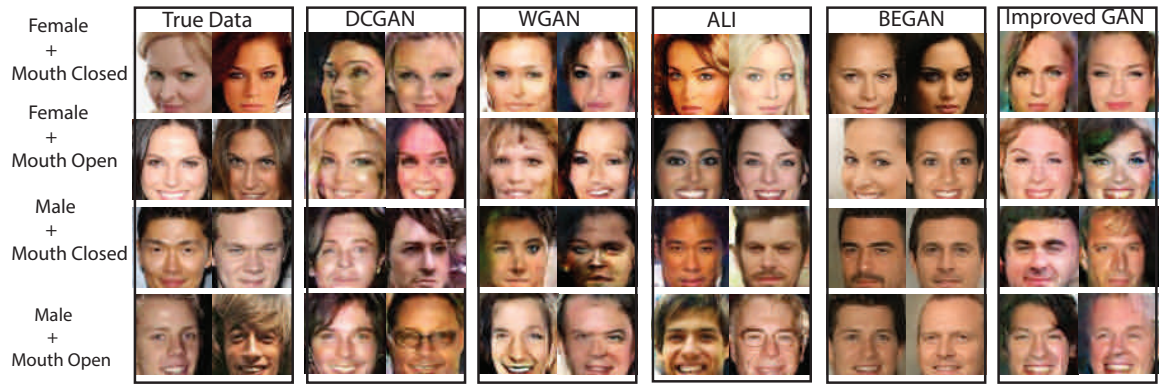

(b) 4-class dataset from CelebA for attributes *Male, Mouth Open*.

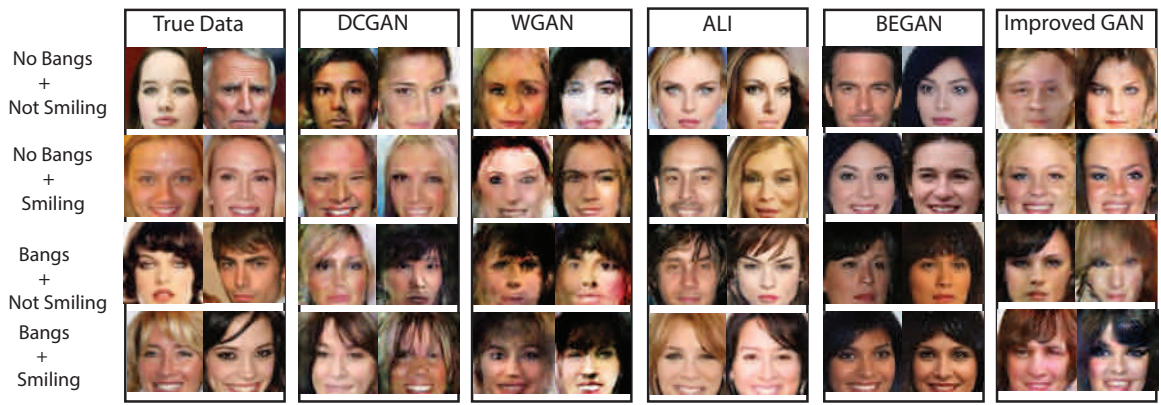

(c) 4-class dataset from CelebA for attributes *Bangs, Smiling*.

*Figure 4.* Illustration of subsets of the CelebA dataset used in our classification-based GAN studies. Shown alongside are samples from various GANs trained on this dataset. Labels for the GAN samples are obtained using a pre-trained classifier as an annotator.

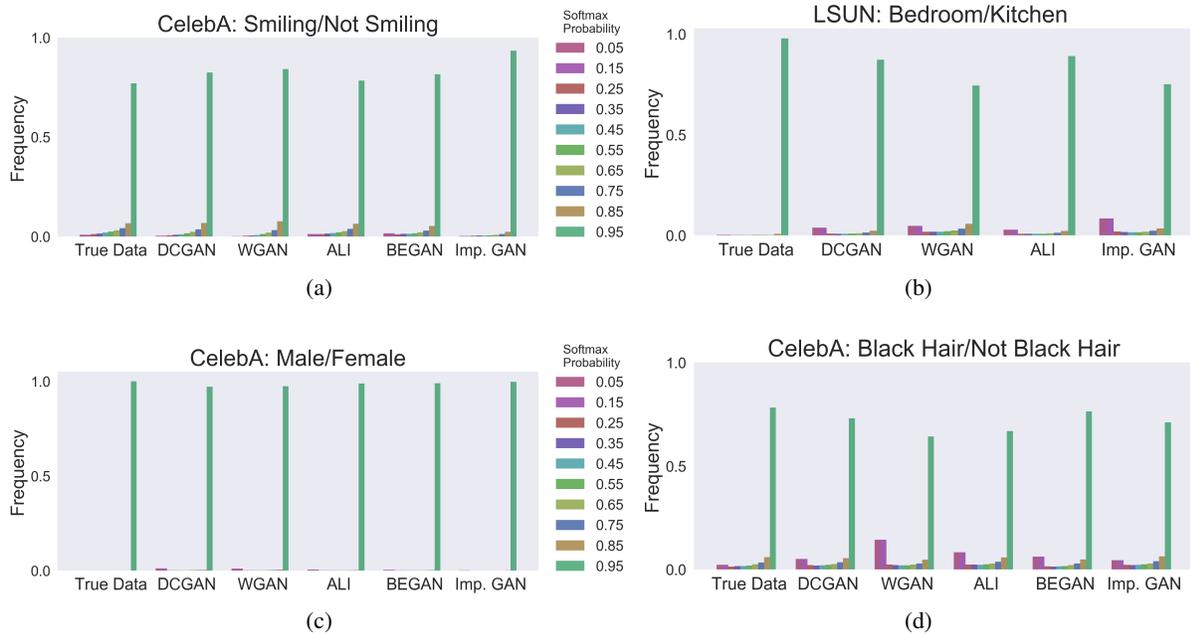

*Figure 5.* Histograms of annotator confidence (softmax probability) during label prediction on true data (test set) and synthetic data for tasks on the CelebA and LSUN datasets (see 5.2).

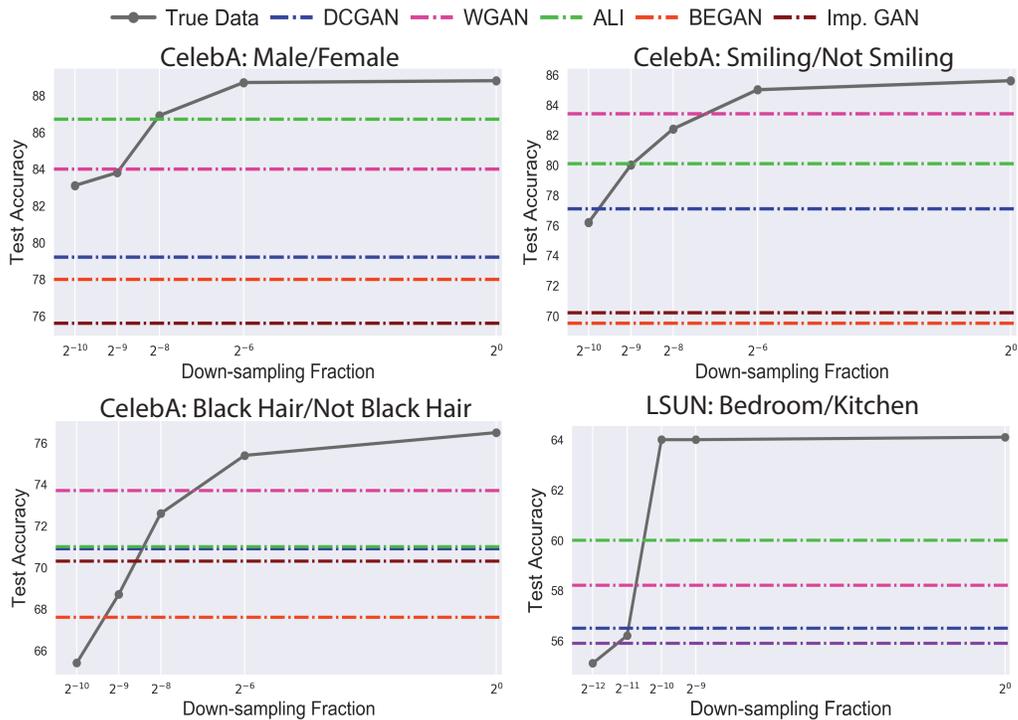

*Figure 6.* Illustration of the classification performance of true data compared with GAN-generated synthetic datasets based on experiments described in 5.2. Classification is performed using a basic linear model, described in 4.2, and performance is reported in terms of accuracy on a hold-out set of true data. In the plots, the bold curve shows the classification performance of models trained on true data vs the size of the true dataset (maximum size is $N$). Dashed lines represent performance of classifiers trained on various GAN-generated datasets of size $N$. These plots indicate that GANs suffer significantly from covariate shift due to boundary distortion.

| Task | Data Source | Classification Performance | | | | | |
|---|---|---|---|---|---|---|---|
| | | Label Correctness (%) | Inception Score ($\mu \pm \sigma$) | Accuracy (%) | | | |
| | | | | Linear model | | | ResNet |
| | | | | $\uparrow_1$ | | $\uparrow_{10}$ | $\uparrow_1$ |
| | | | | Train | Test | Train | Test | Test |
| CelebA Male (Y/N) # Images: 136522 | True | 97.9 | $1.98 \pm 0.0033$ | 88.1 | 88.8 | | | 97.9 |
| | True $\downarrow_{64}$ | | | 89.6 | 88.7 | | | 92.9 |
| | True $\downarrow_{256}$ | | | 91.6 | 86.9 | | | 89.8 |
| | True $\downarrow_{512}$ | | | 96.3 | 83.8 | | | 82.6 |
| | True $\downarrow_{1024}$ | | | 100.0 | 83.1 | | | 81.4 |
| | DCGAN | 98.2 | $1.97 \pm 0.0013$ | 100.0 | 79.2 | 100.0 | 79.6 | 56.4 |
| | WGAN | 98.3 | $1.97 \pm 0.0013$ | 96.7 | 84.0 | 96.7 | 83.9 | 50.0 |
| | ALI | 99.2 | $1.99 \pm 0.0008$ | 95.8 | 86.7 | 95.8 | 86.7 | 58.9 |
| | BEGAN | 99.3 | $1.99 \pm 0.0006$ | 97.9 | 78.0 | 98.0 | 78.2 | 55.4 |
| | Improved GAN | 99.8 | $1.99 \pm 0.0004$ | 100.0 | 75.6 | 100.0 | 71.0 | 71.7 |
| CelebA Smiling (Y/N) # Images: 156160 | True | 92.4 | $1.69 \pm 0.0074$ | 85.7 | 85.6 | | | 92.4 |
| | True $\downarrow_{64}$ | | | 87.6 | 85.0 | | | 87.8 |
| | True $\downarrow_{256}$ | | | 91.5 | 82.4 | | | 82.1 |
| | True $\downarrow_{512}$ | | | 93.7 | 80.0 | | | 77.8 |
| | True $\downarrow_{1024}$ | | | 95.0 | 76.2 | | | 71.2 |
| | DCGAN | 96.1 | $1.67 \pm 0.0028$ | 100.0 | 77.1 | 100.0 | 77.1 | 63.3 |
| | WGAN | 98.2 | $1.68 \pm 0.0031$ | 96.8 | 83.4 | 96.8 | 83.5 | 65.3 |
| | ALI | 93.3 | $1.71 \pm 0.0027$ | 94.5 | 80.1 | 95.0 | 82.4 | 55.8 |
| | BEGAN | 93.5 | $1.74 \pm 0.0028$ | 98.5 | 69.5 | 98.5 | 69.6 | 64.1 |
| | Improved GAN | 98.4 | $1.88 \pm 0.0021$ | 100.0 | 70.2 | 100.0 | 70.1 | 61.6 |
| CelebA Black Hair (Y/N) # Images: 77812 | True | 84.5 | $1.68 \pm 0.0112$ | 76.4 | 76.5 | | | 84.5 |
| | True $\downarrow_{64}$ | | | 79.7 | 75.4 | | | 80.0 |
| | True $\downarrow_{256}$ | | | 86.3 | 72.6 | | | 75.8 |
| | True $\downarrow_{512}$ | | | 89 | 68.7 | | | 73.9 |
| | True $\downarrow_{1024}$ | | | 100.0 | 65.4 | | | 72.7 |
| | DCGAN | 86.7 | $1.68 \pm 0.0040$ | 100.0 | 70.9 | 100.0 | 70.5 | 53.4 |
| | WGAN | 76.0 | $1.60 \pm 0.0055$ | 94.4 | 73.7 | 94.3 | 73.4 | 58.5 |
| | ALI | 79.4 | $1.63 \pm 0.0028$ | 94.9 | 71.0 | 94.9 | 70.2 | 55.7 |
| | BEGAN | 87.6 | $1.74 \pm 0.0028$ | 94.1 | 67.6 | 94.1 | 67.7 | 67.2 |
| | Improved GAN | 86.7 | $1.64 \pm 0.0045$ | 100.0 | 70.3 | 100.0 | 69.1 | 70.2 |
| LSUN Bedroom/Kitchen # Images: 200000 | True | 98.2 | $1.94 \pm 0.0217$ | 64.7 | 64.1 | | | 99.1 |
| | True $\downarrow_{512}$ | | | 64.7 | 64.0 | | | 76.4 |
| | True $\downarrow_{1024}$ | | | 65.2 | 64.0 | | | 66.9 |
| | True $\downarrow_{2048}$ | | | 98.7 | 56.2 | | | 56.5 |
| | True $\downarrow_{4096}$ | | | 100.0 | 55.1 | | | 55.1 |
| | DCGAN | 92.7 | $1.85 \pm 0.0036$ | 90.8 | 56.5 | 91.2 | 56.3 | 51.2 |
| | WGAN | 87.8 | $1.70 \pm 0.0023$ | 86.2 | 58.2 | 96.3 | 54.1 | 55.7 |
| | ALI | 94.1 | $1.86 \pm 0.0021$ | 82.5 | 60.0 | 82.0 | 59.7 | 56.2 |
| | Improved GAN | 84.2 | $1.68 \pm 0.0030$ | 91.6 | 55.9 | 90.8 | 56.5 | 51.2 |

*Table 3.* Detailed results from study on covariate shift between true and GAN distributions on the CelebA and LSUN datasets shown in 1, based on experiments described in 5.2. Label correctness measures the agreement between default labels for the synthetic datasets, and those predicted by the annotator, a classifier trained on the true data. Shown alongside are the modified inception scores computed using labels predicted by the annotator (instead of the Inception Network). Training and test accuracies for a linear model classifier on the various true and synthetic datasets are reported. Also presented are the corresponding accuracies for a linear model trained on down-sampled true data ($\downarrow_M$) and oversampled synthetic data ($\uparrow_L$). Test accuracy for ResNets trained on these datasets is also shown (training accuracy was always 100%), though it is noticeable that deep networks suffer from issues when trained on synthetic datasets.